\documentclass[a4paper,conference]{IEEEtran}
\usepackage{graphicx}
\usepackage{comment}
\usepackage{amsmath,amssymb} 
\usepackage{color}

\usepackage[left=13.1mm,right=13.1mm,top=19.1mm,bottom=39.9mm]{geometry}

\usepackage{ifthen}
\usepackage{algorithm}
\usepackage{algpseudocode}

\usepackage{pgfplots}
\pgfplotsset{compat=1.15}
\usepackage{tikz}

\usepackage{makecell}

\usepackage{hyperref}

\usepackage{balance}

\usetikzlibrary{external}
\begin{document}
%
\title{FeatureNMS: Non-Maximum Suppression by Learning Feature Embeddings}

\author{\IEEEauthorblockN{Niels Ole Salscheider}
\IEEEauthorblockA{FZI Research Center for Information Technology\\
Haid-und-Neu-Str. 10-14\\
76131 Karlsruhe\\
Germany\\
Email: salscheider@fzi.de}
}


\maketitle

\begin{abstract}
Most state of the art object detectors output multiple detections per object.
The duplicates are removed in a post-processing step called \emph{Non-Maximum Suppression}.
Classical Non-Maximum Suppression has shortcomings in scenes that contain objects with high overlap:
This heuristic assumes that a high overlap between two bounding boxes corresponds to a high probability of one being a duplicate.
We propose FeatureNMS to solve this problem.
FeatureNMS recognizes duplicates not only based on the intersection over union between the bounding boxes, but also based on the difference of feature vectors.
These feature vectors can encode more information like visual appearance.
Our approach outperforms classical NMS and derived approaches and achieves state of the art performance.
\end{abstract}


%
\IEEEpeerreviewmaketitle

\section{Introduction}
Object detection is an important task in a huge variety of applications.
In some of these applications, images can contain a lot of partially overlapping objects.
One example are images of traffic scenes that contain crowds of humans.
This scenario is common in autonomous driving or surveillance scenarios.

Most state of the art object detectors are based on Convolutional Neural Networks (CNN).
There are single-stage detectors like YOLO \cite{Redmon2016,Redmon2017,Redmon2018}, SSD \cite{Liu2016} and RetinaNet \cite{Lin2017}, and two-stage detectors like R-CNN \cite{Girshick2014}, Fast R-CNN \cite{Girshick2015} and Faster R-CNN \cite{Ren2015}.
Two-stage detectors first generate a set of proposals.
A dedicated second stage then decides which proposals are in fact an object of interest.
Single-stage detectors on the other hand directly perform object detection on the input image.

Both approaches have in common that they usually generate multiple detections per object.
Duplicate detections are then removed in a post-processing step called \emph{Non-Maximum Suppression} (NMS).
The widely used classical approach is a greedy heuristic.
Detections are sorted by their scores in a decreasing order.
Then each detection is checked against all following in the sorted list.
If the \emph{Intersection over Union} (IoU) with one of the following detections is larger than a certain threshold the latter detection is removed.

This heuristic however has shortcomings in crowded scenes because the underlying assumption does not hold.
In these scenes, distinct objects often have a high overlap.
In this paper we propose \emph{FeatureNMS} to solve this problem.
Our approach recognizes duplicates based on their feature embeddings if a definite decision based on the IoU is not possible.

The remainder of this paper is structured as follows:
Section~\ref{sec:relatedwork} presents related work.
Section~\ref{sec:approach} describes our proposed approach to Non-Maximum Suppression.
The general idea is presented in Section~\ref{sec:approach:proposednms} while Section~\ref{sec:approach:CNN} contains details about the necessary modifications to the object detector network.
In Section~\ref{sec:evaluation} we present our evaluation procedure and the results.
Finally, Section~\ref{sec:conclusion} concludes the paper.

\section{Related Work}
\label{sec:relatedwork}
Both NMS and embedding learning have been studied in previous research.
This section presents relevant and related work in these fields.

\subsection{Non-Maximum Suppression}
There have been several proposals how to improve the classical NMS heuristic.
SoftNMS \cite{Bolda2017} does not remove overlapping detections but decreases the detection scores of duplicates.
The factor by which it is decreased is a function of the IoU of the corresponding bounding boxes.

The idea of AdaptiveNMS \cite{Liu2019} is to adjust the threshold for the greedy heuristic based on the local object density.
This local object density is predicted by the object detection network for each detection.

Visibility Guided NMS \cite{Gahlert2019} uses another approach.
The detection network outputs two bounding boxes per object.
One bounding box encloses the whole object while the other encloses only the visible part.
Given detections of two different objects, the IoU for the visible parts is usually smaller than the IoU for the whole objects.
Because of that, classical NMS is performed on the bounding boxes of the visible parts.
But the final output are the corresponding bounding boxes of the whole object.

Other works \cite{Wang2018,Zhang2018} try to work around the shortcomings of classical NMS during the training of the object detector.
The idea is to push bounding boxes of different objects far enough apart.
Boxes that to the same object, however, should have as much overlap as possible.
This makes the task of NMS easier since the detections violate the underlying assumptions less.

In \cite{Hosang2017}, the authors propose to solve the NMS task with a CNN.
The proposed network learns to re-score detections to suppress duplicates.
Each block in the network has access to pairwise features of detections.
These features include the IoU of both bounding boxes, normalized distances, as well as scale and aspect ratio differences.

Relation Networks \cite{Hu2018} add a relation module to the detection network.
This relation module learns to perform NMS inside the network.
It can use geometric and appearance features of the detections for this.

\subsection{Embedding Learning}
Learning of embeddings is used in a wide range of applications like zero-shot learning \cite{Bucher2016}, visual search \cite{Wang2014,Bala2015,Shankar2017} or image comparison \cite{Hoffer2015,Dey2017}.
The underlying idea is conceptionally simple:
The embedding vectors of positive image pairs (i.\,e. images that show the same object) should be similar.
Embedding vectors of negative pairs on the other hand should be separated by a certain distance.

There are several loss functions that can be used to achieve this objective.
Contrastive loss \cite{Hadsell2006} is widely used for this purpose.
It consists of two terms:
One term pulls the $\ell^2$ distance of positive pairs as close to zero as possible.
The other term pushes the $\ell^2$ distance of negative pairs apart if it is below a certain margin.

Choosing the margin parameter correctly can be challenging.
It can be too difficult to push the embeddings of hard negative examples far enough apart while keeping small distances for positive pairs.
Triplet loss \cite{Schroff2015} tries to solve this problem by using triplets of images: An anchor, a positive example and a negative example.
It tries to ensure that the embedding of the anchor is closer to all positive examples than to any negative example.
The authors also propose a sampling strategy to select suitable triplets for training.

Recently, Margin loss \cite{Wu2017} has been proposed as an alternative to contrastive loss.
It does not try to push the embeddings of all positive pairs to be as close to each other as possible.
Instead, it just requires the distance to be below a certain margin, making the loss more robust.
Together with a distance weighted sampling strategy it achieves state of the art performance on multiple tasks.

\section{Approach}
\label{sec:approach}

We first describe our proposed approach for Non-Maximum Suppression in Section~\ref{sec:approach:proposednms}.
We then describe the necessary modifications to the object detector and the training procedure in Section~\ref{sec:approach:CNN}.

\subsection{Proposed Non-Maximum Suppression}
\label{sec:approach:proposednms}

With classical Non-Maximum Suppression, all detections are first sorted by their confidence scores and added to a proposal list $\mathcal{P}$.
The list of final detections $\mathcal{D}$ is empty in the beginning.
Then the following step is executed iteratively until $\mathcal{P}$ is empty:
The proposal $p$ with the highest confidence score in $\mathcal{P}$ is removed from $\mathcal{P}$ and compared to all detections in $\mathcal{D}$.
If the intersection over union between $p$ and all detections in $\mathcal{D}$ is smaller than a threshold $N$ then $p$ is added to $\mathcal{D}$.
Otherwise $p$ is discarded.
The pseudo code of this algorithm is given in Algorithm~\ref{algo:classicalnms}.

\begin{algorithm}
 \begin{algorithmic}
  \State $\mathcal{P} \gets \Call{getProposals}{image}$
  \State $\mathcal{P} \gets \Call{sort}{\mathcal{P}}$
  \State $\mathcal{D} \gets \emptyset$

  \While{$\mathcal{P} \ne \emptyset$}
   \State $p \gets \Call{pop}{\mathcal{P}}$
   \State $isDuplicate \gets \text{false}$
   \For{$d \in \mathcal{D}$}
    \State $iou \gets \Call{getIoU}{p, d}$
    \If{$iou > N$}
     \State $isDuplicate \gets \text{true}$
    \EndIf
   \EndFor
   \If{$\neg isDuplicate$}
    \State $\Call{push}{p, \mathcal{D}}$
   \EndIf
  \EndWhile
 \end{algorithmic}
 \caption{Classical Non-Maximum Suppression.}
 \label{algo:classicalnms}
\end{algorithm}

This approach has one parameter $N$ which has to be tuned to achieve good performance.
A common choice is $N = 0.5$.
The key idea of this algorithm is that bounding boxes with a high overlap are likely to belong to the same object.
Bounding boxes with a low overlap on the other hand are likely to belong to different objects.

There are however situations where this assumption fails.
Especially in images with a high number of objects and partial occlusions there are many overlapping bounding boxes that belong to different objects.
One example of this are crowds of humans.

We propose a novel approach to decide if two bounding boxes belong to the same object or not.
We call our approach FeatureNMS since it is based on (appearance) features of the detections.
The overall structure of the proposed algorithm is the same as classical Non-Maximum Suppression---but the rule whether to add $p$ from $\mathcal{P}$ to $\mathcal{D}$ or not is adjusted.
The pseudo code of our approach is given in Algorithm~\ref{algo:proposednms}.

\begin{algorithm}
 \begin{algorithmic}
  \State $\mathcal{P} \gets \Call{getProposals}{image}$
  \State $\mathcal{P} \gets \Call{sort}{\mathcal{P}}$
  \State $\mathcal{D} \gets \emptyset$

  \While{$\mathcal{P} \ne \emptyset$}
   \State $p \gets \Call{pop}{\mathcal{P}}$
   \State $isDuplicate \gets \text{false}$
   \For{$d \in \mathcal{D}$}
    \State $iou \gets \Call{getIoU}{p, d}$
    \If{$iou > N_2$}
     \State $isDuplicate \gets \text{true}$
    \ElsIf{$iou > N_1$}
     \State $embeddingDist \gets \Call{getEmbeddingDist}{p, d}$
     \If{$embeddingDist < T$}
      \State $isDuplicate \gets \text{true}$
     \EndIf
    \EndIf
   \EndFor
   \If{$\neg isDuplicate$}
    \State $\Call{push}{p, \mathcal{D}}$
   \EndIf
  \EndWhile
 \end{algorithmic}
 \caption{Proposed Non-Maximum Suppression.
  If the calculated value of the intersection over union is in a range that does not allow to make a definite decision we use a feature embedding similarity.}
 \label{algo:proposednms}
\end{algorithm}

Again, each proposal $p \in \mathcal{P}$ is compared to all detections $d \in \mathcal{D}$.
The intersection over union between $p$ and $d$ is computed.
If this value is less or equal than a threshold $N_1$ we assume that the detections belong to different objects.
If this value on the other hand is larger than another threshold $N_2$ we assume that the detections must belong to the same object.
In any other case the two bounding boxes might belong to the same or to different objects---the intersection over union alone cannot be used to make a final decision.
In this case we calculate the $\ell^2$ distance of the feature embeddings of both bounding boxes.
If this distance is larger than a threshold $T$ we assume that the bounding boxes belong to different objects.
Otherwise they are likely to belong to the same object.
The feature embeddings are an output of the CNN that we use for object detection.
It is described in detail in Section~\ref{sec:approach:CNN}.

We propose to choose $N_1 = 0.1$ and $N_2 = 0.9$ but other values are possible, depending on the application.
The right value for $T$ depends on the training objective of the detection network.
In our work we use $T = \beta = 1.0$ (cf. Section~\ref{sec:approach:CNN}).

\subsection{Detector Architecture and Training}
\label{sec:approach:CNN}
We evaluate our approach with the RetinaNet \cite{Lin2017} object detector.
But it generalizes to many other detector architectures---the only required change is to learn an embedding vector per detection.
For this, we add one network head to each output pyramid level of the RetinaNet backbone.
The head outputs an embedding vector for each anchor box.
We chose an embedding of length of 32, but other lengths are possible.
In our experiments, this proved to be  a good trade-off between accuracy, computational overhead and memory consumption.

Like all other RetinaNet heads, the network head for the feature embeddings consists of four identical blocks.
Each block is formed by a 2D convolution layer with 512 channels, a Batch Normalization \cite{Ioffe2015} layer and a ReLU activation function.
The output of the last block is $\ell^2$-normalized along the embedding dimension (consisting of 32 values).
This ensures that all embeddings lie on a unit hypersphere which is a common choice for embedding learning \cite{Schroff2015}.

The training objective for the feature embedding is based on Margin Loss \cite{Wu2017}.
The total loss can be calculated as follows:

\begin{equation}
  L = \frac{\sum_{i \in \mathcal{A}} \sum_{j \in \mathcal{A} \setminus \{i\}} L'(i, j)}{\lvert \mathcal{A} \rvert \cdot (\lvert \mathcal{A} \rvert - 1)}
\end{equation}

In this equation $L'$ is the pairwise loss between two targets:

\begin{equation}
 L'(i, j) =
 \begin{cases}
  \max\left(0, \left\lVert \mathbf{f}_i, \mathbf{f}_j \right\rVert_2 - (\beta - \alpha)\right), & \text{if } obj(i) = obj(j) \\
  \max\left(0, (\beta + \alpha) - \left\lVert \mathbf{f}_i, \mathbf{f}_j \right\rVert_2\right), & \text{otherwise}
 \end{cases}
\end{equation}

Here, $\mathcal{A}$ is the set of anchor boxes that are assigned to ground truth bounding boxes.
The vector $\mathbf{f}_i$ is the embedding feature vector that belongs to the target (anchor box) $i$.
The function $obj(i)$ gives the object id of target $i$.
The parameter $\alpha$ determines the margin between positive and negative examples, and the parameter $\beta$ determines the decision threshold.
We chose $\alpha = 0.2$ and $\beta = 1.0$.

Our sampling strategy is different from \cite{Wu2017}.
Since we only train on active target pairs within a single image, the number of pairs is limited.
This means that we can nearly always use all possible pairs during a training step.
Only if the number of pairs exceeds 5\,000, we use uniform sampling to restrict the number of samples to 5\,000.

We weight the different losses during training according to \cite{Kendall2017}.
This way, the weighting factors can adjust based on the training progress and do not have to be tuned manually.

\section{Evaluation}
\label{sec:evaluation}

We evaluate our approach on the CrowdHuman dataset \cite{Shao2018}.
This dataset contains 15\,000 training images and 4\,370 validation images.
We use the validation images to compare the performance of the different NMS approaches, but we did not use it to tune any parameters.
The dataset contains multiple annotations per person: A head bounding box, a visible region bounding box and a full body bounding box.
In this work, we use the visible body bounding box annotations.
Before feeding the images into the network, we resize them so that the longer side has a fixed amount of pixels.
Then the image is padded with a fixed color value to obtain a square image.

Our implementation is based on the RetinaNet implementation from Tensorflow\footnote{\url{https://github.com/tensorflow/models}}.
Our patches for this implementation that we used to perform the experiments are available online\footnote{\url{https://github.com/fzi-forschungszentrum-informatik/NNAD/tree/featurenms}}.

We use the default hyperparameters with the following exceptions:
\begin{itemize}
 \item Batch size of 4
 \item 800\,000 training steps
 \item LAMB optimizer \cite{You2019}
 \item Learning rate
 \begin{itemize}
  \item $1 \cdot 10^{-4}$ (step 0 - 100\,000)
  \item $5 \cdot 10^{-5}$ (step 100\,000 - 200\,000)
  \item $1 \cdot 10^{-5}$ (step 200\,000 - 400\,000)
  \item $5 \cdot 10^{-6}$ (step 400\,000 - 800\,000)
 \end{itemize}
 \item Image size
 \begin{itemize}
  \item $768 \times 768$ pixels (first 750\,000 training steps)
  \item $1024 \times 1024$ pixels (last 50\,000 training steps and during testing)
 \end{itemize}
\end{itemize}

We initialized the weights of our CNN backbone from a model that was pretrained on the COCO dataset \cite{Lin2014}.
During training, we froze the weights of the first convolutional layer and the corresponding batch normalization layer.

Most of the training steps were performed at a reduced resolution of $768 \times 768$ pixels.
The reason is that the limited VRAM of our GPU does not allow to train at higher resolutions with a batch size of 4.
Afterwards we fine-tuned the network at full resolution on the CPU for $50\,000$ training steps.

We evaluate the different NMS approaches with three common metrics.
The first is the average precision when requiring an IoU of at least 0.5 between detection and ground truth bounding box.
The second is the average precision at a minimum IoU of 0.75.
The last metric that we use is the log-average miss rate \cite{Dollar2012}.
This metric is computed by averaging miss rates at 9 FPPI (false positives per image) values evenly spaced in log-space between $10^{-2}$ and $10^0$.
The IoU threshold used for this is 0.5.

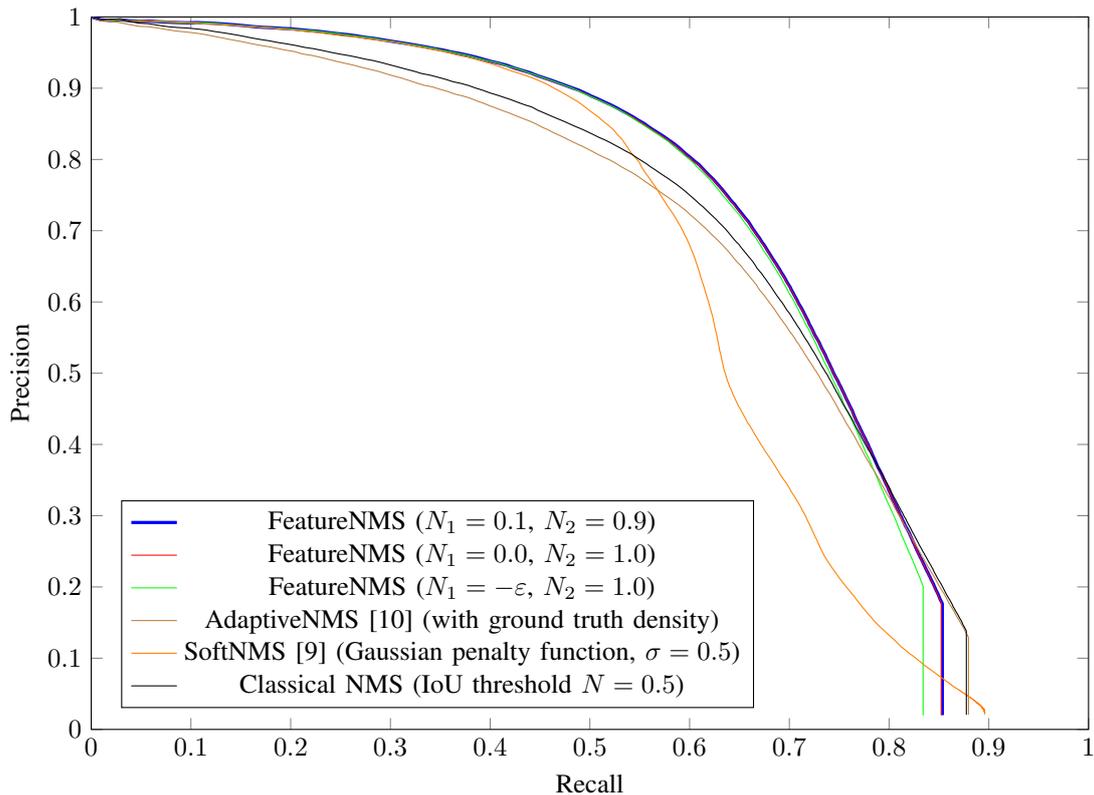
\begin{figure*}
 \begin{center}
 \begin{tikzpicture}
  \begin{axis}[xlabel=Recall, ylabel=Precision, xmin=0, xmax=1, ymin=0, ymax=1, width=0.8\textwidth, height=0.6\textwidth, legend pos=south west]
   \addplot [blue, very thick, no markers] gnuplot [raw gnuplot] { plot "content/pr_curve_0.1_0.9.csv" every 1000; };
   \addlegendentry{FeatureNMS ($N_1 = 0.1$, $N_2 = 0.9$)}
   \addplot [red, no markers] gnuplot [raw gnuplot] { plot "content/pr_curve_0.0_1.0.csv" every 1000; };
   \addlegendentry{FeatureNMS ($N_1 = 0.0$, $N_2 = 1.0$)}
   \addplot [green, no markers] gnuplot [raw gnuplot] { plot "content/pr_curve_-eps_1.0.csv" every 1000; };
   \addlegendentry{FeatureNMS ($N_1 = -\varepsilon$, $N_2 = 1.0$)}
   \addplot [brown, no markers] gnuplot [raw gnuplot] { plot "content/pr_curve_adaptivenms.csv" every 1000; };
   \addlegendentry{AdaptiveNMS \cite{Liu2019} (with ground truth density)}
   \addplot [orange, no markers] gnuplot [raw gnuplot] { plot "content/pr_curve_softnms.csv" every 1000; };
   \addlegendentry{SoftNMS \cite{Bolda2017} (Gaussian penalty function, $\sigma = 0.5$)}
   \addplot [black, no markers] gnuplot [raw gnuplot] { plot "content/pr_curve_standard_nms.csv" every 1000; };
   \addlegendentry{Classical NMS (IoU threshold $N = 0.5$)}
  \end{axis}
 \end{tikzpicture}
 \end{center}
 \caption{Precision-Recall curves of different approaches for NMS on the CrowdHuman dataset \cite{Shao2018}.
 }
 \label{figure:prcurves}
\end{figure*}

The results can be found in Table~\ref{table:nmsresults}.
We also provide precision-recall curves for all approaches in Figure~\ref{figure:prcurves}.
All reported values are based on the output of the same detector network---only the NMS approach differs.

Our approach (FeatureNMS, $N_1 = 0.1$, $N_2 = 0.9$) outperforms all other approaches that we compared to.
As an ablation study, we evaluated our approach with different parameters and found that the performance does not change much.
When using $N_1 = 0.0$ and $N_2 = 1.0$ the only assumption is that bounding boxes without any overlap can't belong to the same object.
If there is any overlap the feature vector is always used to make the final decision.
When using $N_1 = -\varepsilon$ and $N_2 = 1.0$, even this assumption is given up.
For each pair of detections in an image, the feature vector is used to decide if a box should be suppressed.
This experiment shows the discriminativeness of our feature vector.
Even with these parameters, precision and recall are high and our approach still performs better than the others.

The performance of classical NMS is below that of FeatureNMS except for very low detection score thresholds.
Here, the precision is low for both approaches but the recall of classical NMS is slightly higher.
This is because in a few cases the feature vectors of detections that belong to different objects are too similar.
These detections are erroneously suppressed by FeatureNMS, but not by classical NMS.

SoftNMS \cite{Bolda2017} achieves similar precision as FeatureNMS at high detection score thresholds with low recall.
But the precision at higher recall values is much lower.

We also compared our approach to AdaptiveNMS \cite{Liu2019}.
AdaptiveNMS predicts the local object density for each detection and uses that to adjust the threshold of classical NMS.
We did not want to adjust the detector network for this because a bad network design or training approach could distort the achieved accuracy:
If the density estimation by the detector is not accurate it could reduce the performance of AdaptiveNMS.
Because of that we decided to use the ground truth density as input to AdaptiveNMS.
This also means that the density estimation performance is an overestimate---a real-world detector will not achieve a perfect estimation.

To our surprise we found that AdaptiveNMS performs slightly worse than classical NMS with this ground truth density.
The precision is slightly below that of classical NMS on nearly all points of the precision-recall curve.
This is because the threshold for NMS is increased in densely populated regions of the image, which also leads to more false positives in these regions.
Our findings are in contrast to the results reported in \cite{Liu2019}.
There are several possible explanations for this:
One is that the localization performance of our detector is lower than that of the detectors used in the original paper.
A lower localization performance will result in more false positives when the NMS threshold is high.
Another possible explanation is that the ground truth density is actually not the best threshold:
The neural network might not output a good density estimation, but a smoothed estimation that is closer to the average.
This could suppress some false positives in areas with high object densities.

\begin{figure*}
 \begin{center}
 \begin{tabular}{c c}
  FeatureNMS & Classical NMS \\
  \includegraphics[width=0.4\linewidth]{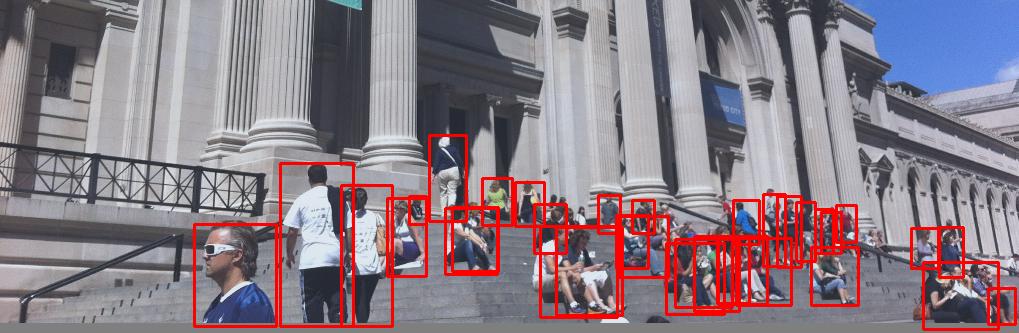} & \includegraphics[width=0.4\linewidth]{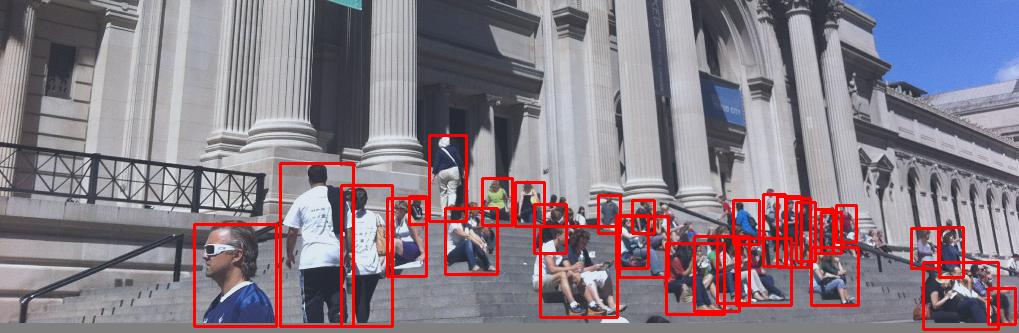} \\
  \includegraphics[width=0.4\linewidth]{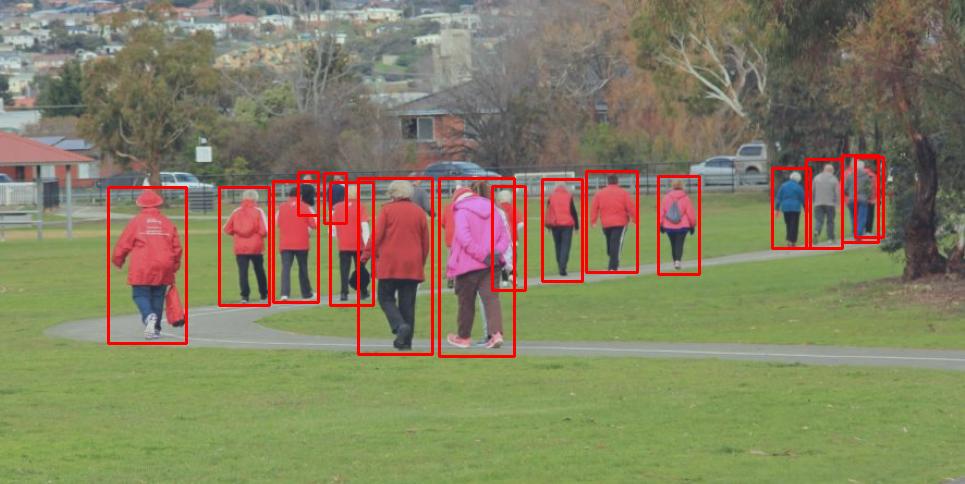} & \includegraphics[width=0.4\linewidth]{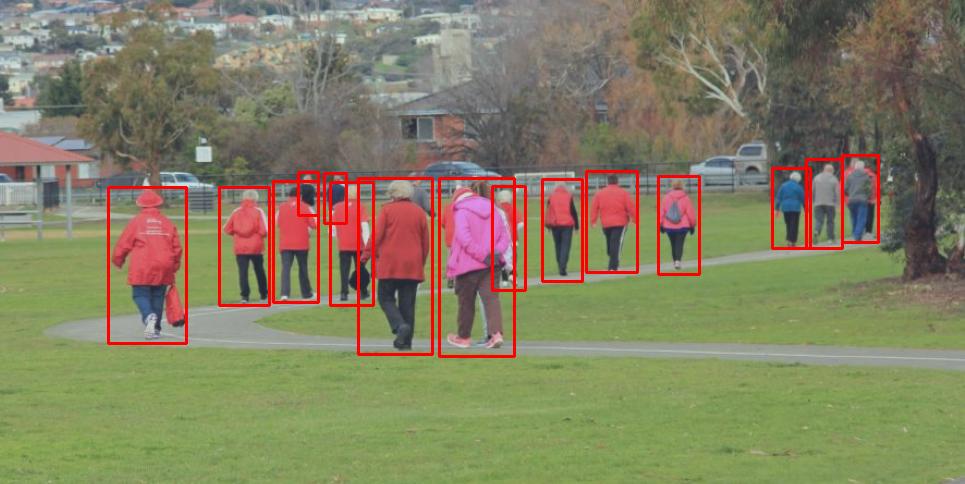} \\
  \includegraphics[height=0.4\linewidth]{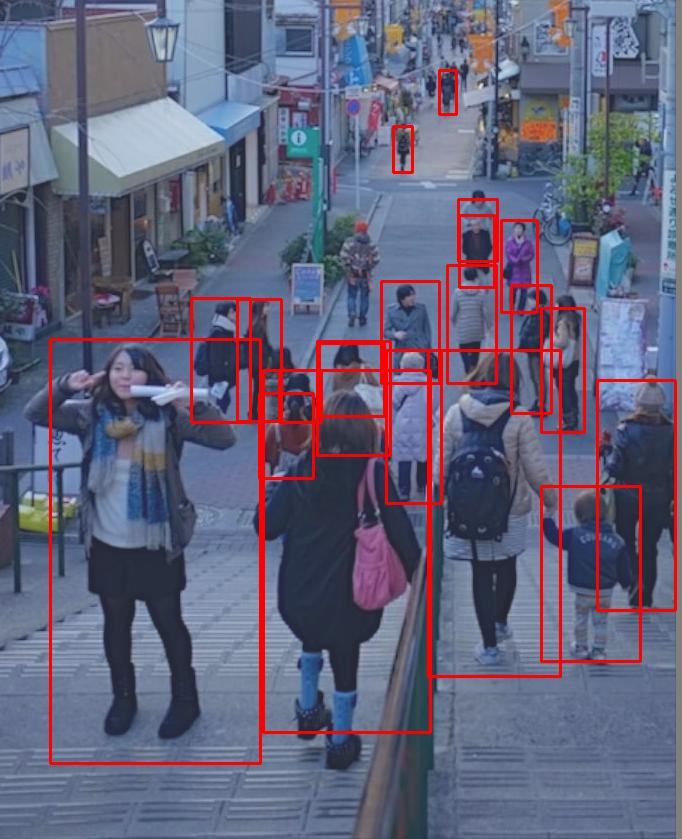} & \includegraphics[height=0.4\linewidth]{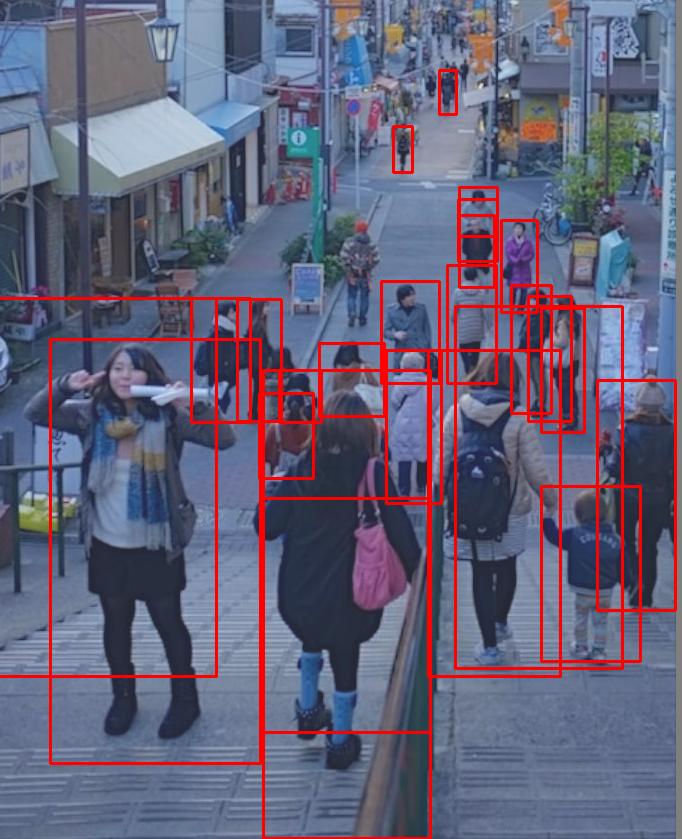} \\
  \includegraphics[height=0.4\linewidth]{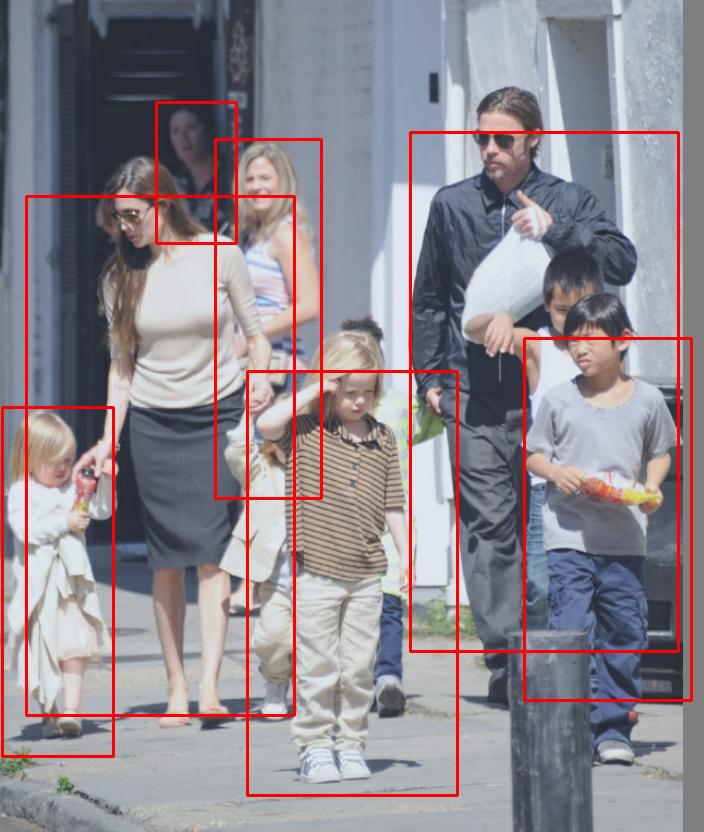} & \includegraphics[height=0.4\linewidth]{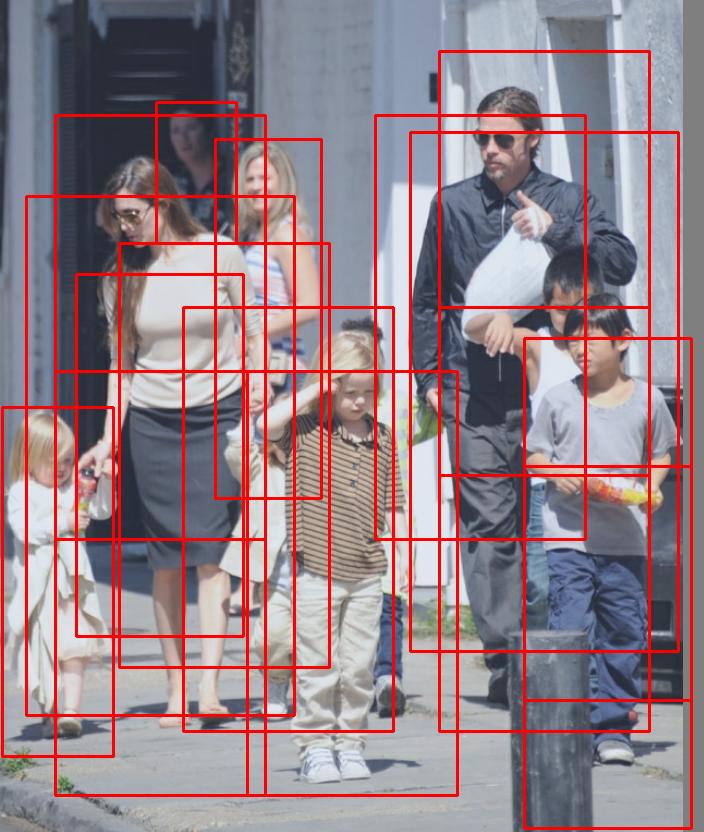}
 \end{tabular}
 \end{center}

 \caption{Comparison of example images when applying FeatureNMS and classical NMS.}
 \label{figure:example_imgs} 
\end{figure*}

We also visually compared the detection results of our approach to these of classical NMS.
Figure~\ref{figure:example_imgs} contains some example images.
We found that there are two situations where FeatureNMS outperforms classical NMS.
The first situation occurs in the first two example images.
Here, there are detections with high overlap that belong to different objects.
Classical NMS suppresses some of these detections while FeatureNMS can correctly separate these.
The second situation occurs in the second two example images.
Here, the bounding box detector outputs some detections with low localization accuracy.
Because of that, the IoU between multiple detections for the same object is low.
Classical NMS fails to suppress the duplicates.
FeatureNMS on the other hand is still able to correctly associate the detections based on the feature vector.

\begin{table}
 \begin{center}
 \begin{tabular}{l | c | c | c}
  Method & AP @ 0.5IoU & AP @ 0.75IoU & \makecell[l]{log-average \\ MR} \\
  \hline
  \hline
  \makecell[l]{FeatureNMS \\ ($N_1 = 0.1$, \\ $N_2 = 0.9$)} & \textbf{0.6865} & \textbf{0.3030} & \textbf{0.7535} \\
  \hline
  \makecell[l]{FeatureNMS \\ ($N_1 = 0.0$, \\ $N_2 = 1.0$)} & 0.6860 & 0.3027 & 0.7545 \\
  \hline
  \makecell[l]{FeatureNMS \\ ($N_1 = -\varepsilon$, \\ $N_2 = 1.0$)} & 0.6838 & 0.2996 & 0.7541 \\
  \hline
  \makecell[l]{AdaptiveNMS \cite{Liu2019} \\ (with ground \\ truth density)} & 0.6480 & 0.2843 & 0.8309 \\
  \hline
  \makecell[l]{SoftNMS \cite{Bolda2017} \\ (Gaussian, \\ $\sigma = 0.5$)} & 0.6280 & 0.2991 & 0.7582  \\
  \hline
  \makecell[l]{Classical NMS \\ (IoU threshold \\ $N = 0.5$)} & 0.6597 & 0.2855 & 0.8129
 \end{tabular}
 \end{center}
 \caption{Comparison of different approaches for NMS on the CrowdHuman dataset \cite{Shao2018}.
 We evaluated the average precision (AP) at a minimum IoU of 0.5 and 0.75, as well as the log-average miss rate (MR) \cite{Dollar2012}.
 Our approach (FeatureNMS) outperforms all other approaches used for comparison.}
 \label{table:nmsresults}
\end{table}

\section{Conclusion}
\label{sec:conclusion}
FeatureNMS is a simple yet effective approach to Non-Maximum Suppression.
It outperforms all approaches that we used for comparison on the CrowdHuman dataset \cite{Shao2018}.
At the same time, the run-time overhead during inference is low:
It performs the same operations as classical NMS.
Additionally to these, it only requires to compute a feature vector per bounding box detection and to compare them for overlapping bounding boxes.
The necessary changes in the object detector network are minor and the approach can be used with most CNN detector architectures.

\balance
\bibliographystyle{IEEEtran}
\bibliography{content/mybibliography}

\end{document}